\pdfoutput=1

\documentclass[11pt]{article}

\usepackage[]{acl}

\usepackage{times}
\usepackage{latexsym}

\usepackage[T1]{fontenc}

\usepackage[utf8]{inputenc}

\usepackage{microtype}
\usepackage{graphicx}
\usepackage{CJKutf8}
\usepackage{booktabs}
\usepackage{multirow}
\usepackage{enumitem}

\title{Kaggle Competition: Cantonese Audio-Visual Speech Recognition for In-car Commands}

\author{Hong Kong University of Science and Technology}

\author{Wenliang Dai, Samuel Cahyawijaya, Tiezheng Yu, Elham J Barezi, Pascale Fung \\
Center for Artificial Intelligence Research (CAiRE)\\
Department of Electronic and Computer Engineering\\
The Hong Kong University of Science and Technology, Clear Water Bay, Hong Kong\\
\{wdaiai,scahyawijaya,tyuah\}@connect.ust.hk, \{eebarezi, pascale\}@ece.ust.hk}

\begin{document}
\setlist{itemsep=5pt,parsep=0pt,topsep=0pt,partopsep=0pt}

\maketitle

\section{Overview}
With the rise of deep learning and intelligent vehicles, the smart assistant has become an essential in-car component to facilitate driving and provide extra functionalities. In-car smart assistants should be able to process general as well as car-related commands and perform corresponding actions, which eases driving and improves safety. However, in this research field, most datasets are in major languages~\citep{dai-etal-2020-kungfupanda,Lovenia2021ASCENDAS,Yu2022AutomaticSR}, such as English and Chinese. There is a huge data scarcity issue for low-resource languages, hindering the development of research and applications for broader communities. Therefore, it is crucial to have more benchmarks 
to raise awareness and motivate the research in low-resource languages.

To mitigate this problem, we collect a new dataset, namely \textbf{C}antonese \textbf{I}n-car \textbf{A}udio-\textbf{V}isual \textbf{S}peech \textbf{R}ecognition (CI-AVSR)~\citep{Dai2022CIAVSRAC}, for in-car speech recognition in the Cantonese language with video and audio data (more details in Section~\ref{sec:dataset}). Together with it, we propose \textbf{\textit{Cantonese Audio-Visual Speech Recognition for In-car Commands}} as a new challenge for the community to tackle low-resource speech recognition under in-car scenarios. 

We expect this challenge to have two major impacts to the community: 
\begin{itemize}
    
    \item As the first Cantonese in-car speech recognition dataset, this challenge paves the way for other low-resource in-car speech recognition and spark enthusiasm of researchers to strenghen the research works in this domain.
    \item The outcome of this challenge will be valuable for social good since Cantonese is a widely spoken language with more than 80M speakers worldwide
    ~\footnote{\url{https://ethnologue.com/language/yue}}.
\end{itemize}

\section{CI-AVSR Challenge}

\subsection{Problem Setup}

The CI-AVSR challenge aims to address typical smart driving scenarios where the driver commands the in-car virtual assistant to perform actions to complete tasks, which can ease the difficulty and improve safety of driving. We provide both video and audio data allowing the model to understand multimodal information to better recognize in-car speech commands in Cantonese. 
Moreover, model robustness is the key to improve the user experience of in-car smart assistants. For this reason, CI-AVSR simulates real in-car environments, which could be very complex, and covers multiple in-car scenarios including both clean and noisy in-car speech recognition scenarios.


\begin{table}[t]
\centering
\resizebox{\linewidth}{!}{
\begin{tabular}{l|ccc}
\toprule
Split & \#Male (Dur.) & \#Female (Dur.) & Total Dur. \\ \midrule \midrule
Train & 10 (10,803s)   & 10 (11,813s)     & 22,616s     \\
Valid & 2 (1,849s)     & 2 (1,829s)       & 3,678s      \\
Test  & 3 (1,902s)     & 3 (1,843s)       & 3,745s      \\ \bottomrule
\end{tabular}
}
\caption{Statistics of the train/valid/test splits of our recorded dataset. We split data by the speaker id, i.e. speakers in a set will not appear in another. Here, \textbf{Dur.} denotes duration and \textbf{s} represents seconds.}
\label{tab:data_split}
\end{table}

\begin{figure*}[t!]
    \centering
    \includegraphics[width=0.90\linewidth]{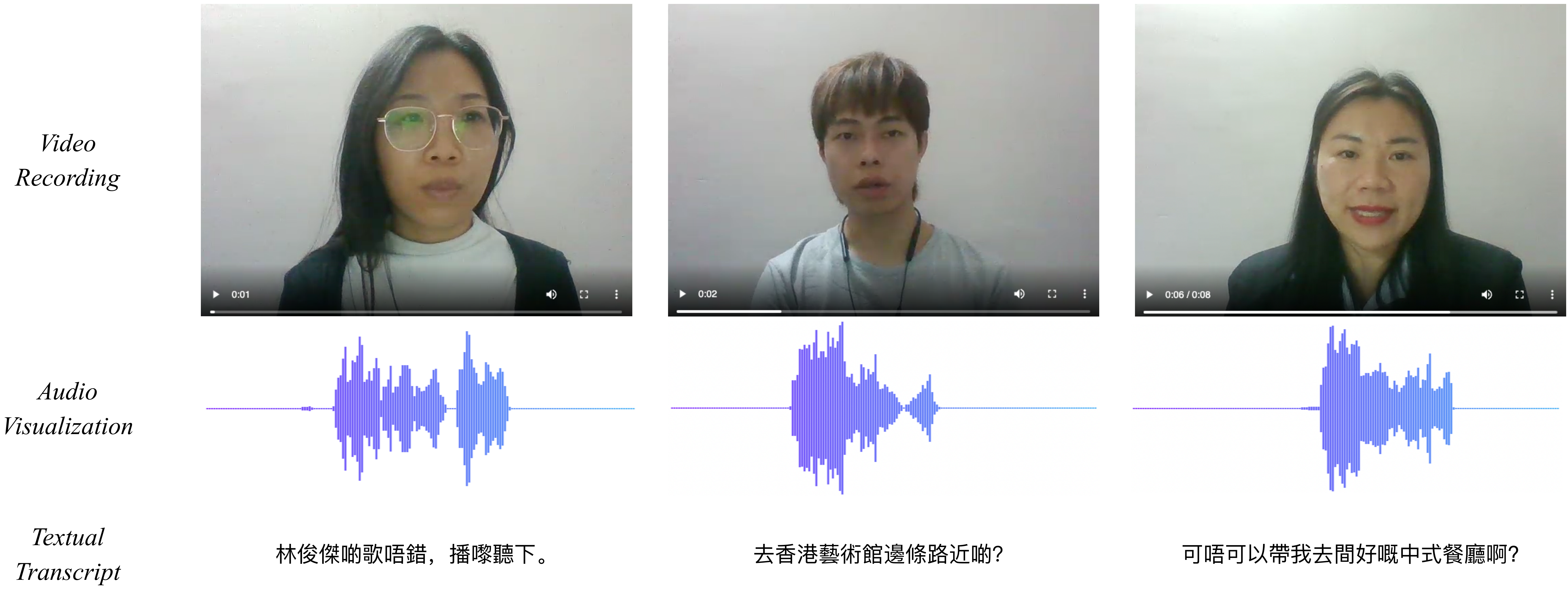}
    \caption{Examples of the CI-AVSR dataset. Each sample in the CI-AVSR consists of an audio-visual clip and the corresponding transcript. The task objective is to generate the transcript from the given audio-visual information.}
    \label{fig:examples}
\end{figure*}

\subsection{Dataset Description} \label{sec:dataset}
The CI-AVSR dataset~\citep{Dai2022CIAVSRAC} consists of 4,984 samples (8.3 hours of video and audio) of 200 unique in-car commands recorded by 30 native Cantonese speakers. We collect the data in a multimodal setup as incorporating more modalities is shown to be very effective in many related research areas~\citep{Afouras2018DeepAS,dai-etal-2020-modality,dai-etal-2021-multimodal,Ma2021EndToEndAS,yu-etal-2021-vision,dai-etal-2022-enabling}. Each command falls in one of the four categories: 1) \textit{navigation}; 2) \textit{music playing}; 3) \textit{weather inquiry}; and 4) \textit{others}. To ensure the quality of commands and make them conform to the spoken Cantonese language, two human experts with native Cantonese proficiency are hired to design and evaluate the commands. Command examples of each category are shown in Table~\ref{tab:commands_template}.

\begin{figure}[h!]
    \centering
    \includegraphics[width=0.93\linewidth]{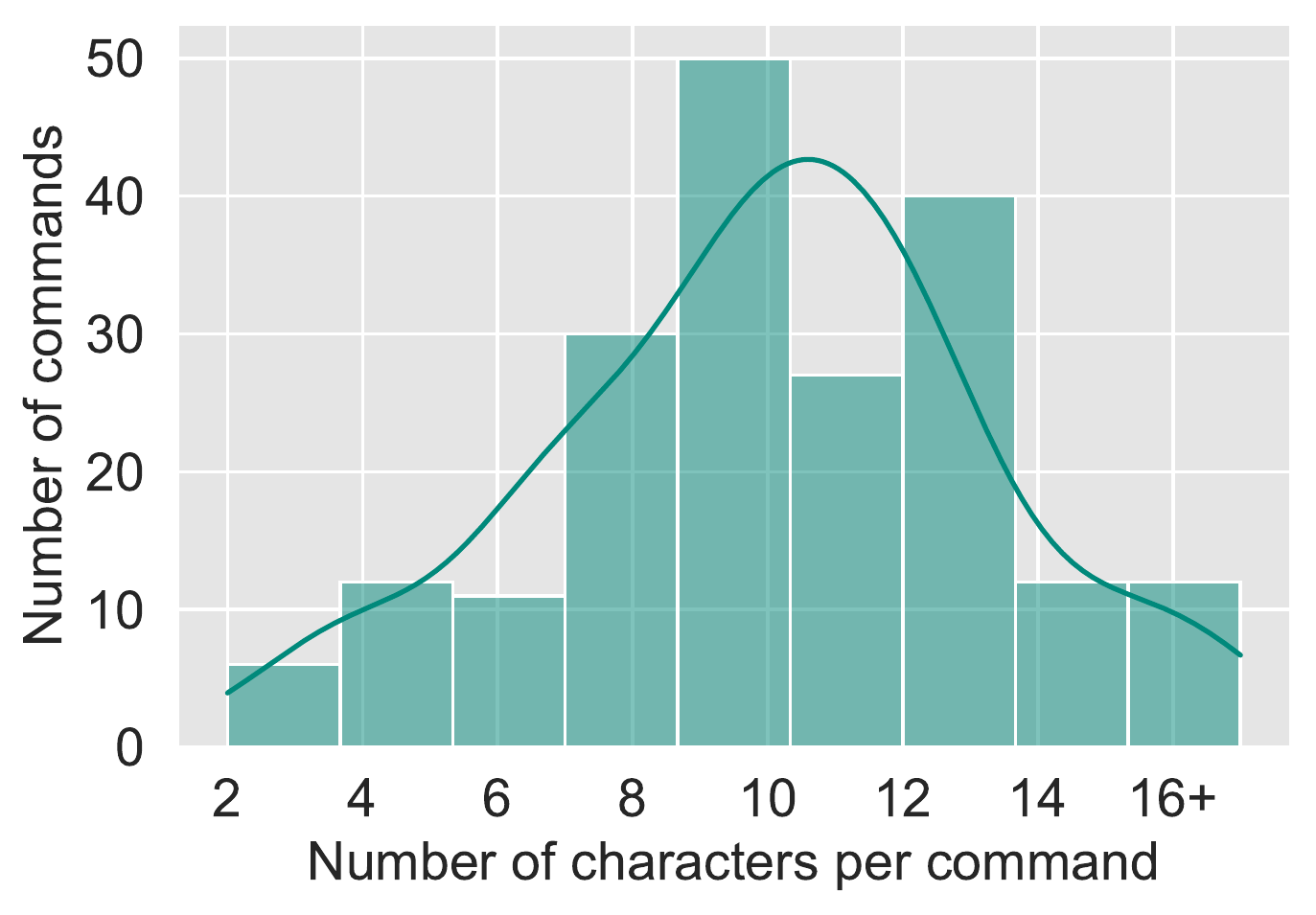}
    \caption{Distribution of the number of characters in all commands. English words are counted as one character. 
    }
    \label{fig:num_chars}
\end{figure}
\begin{figure}[h!]
    \centering
    \includegraphics[width=0.93\linewidth]{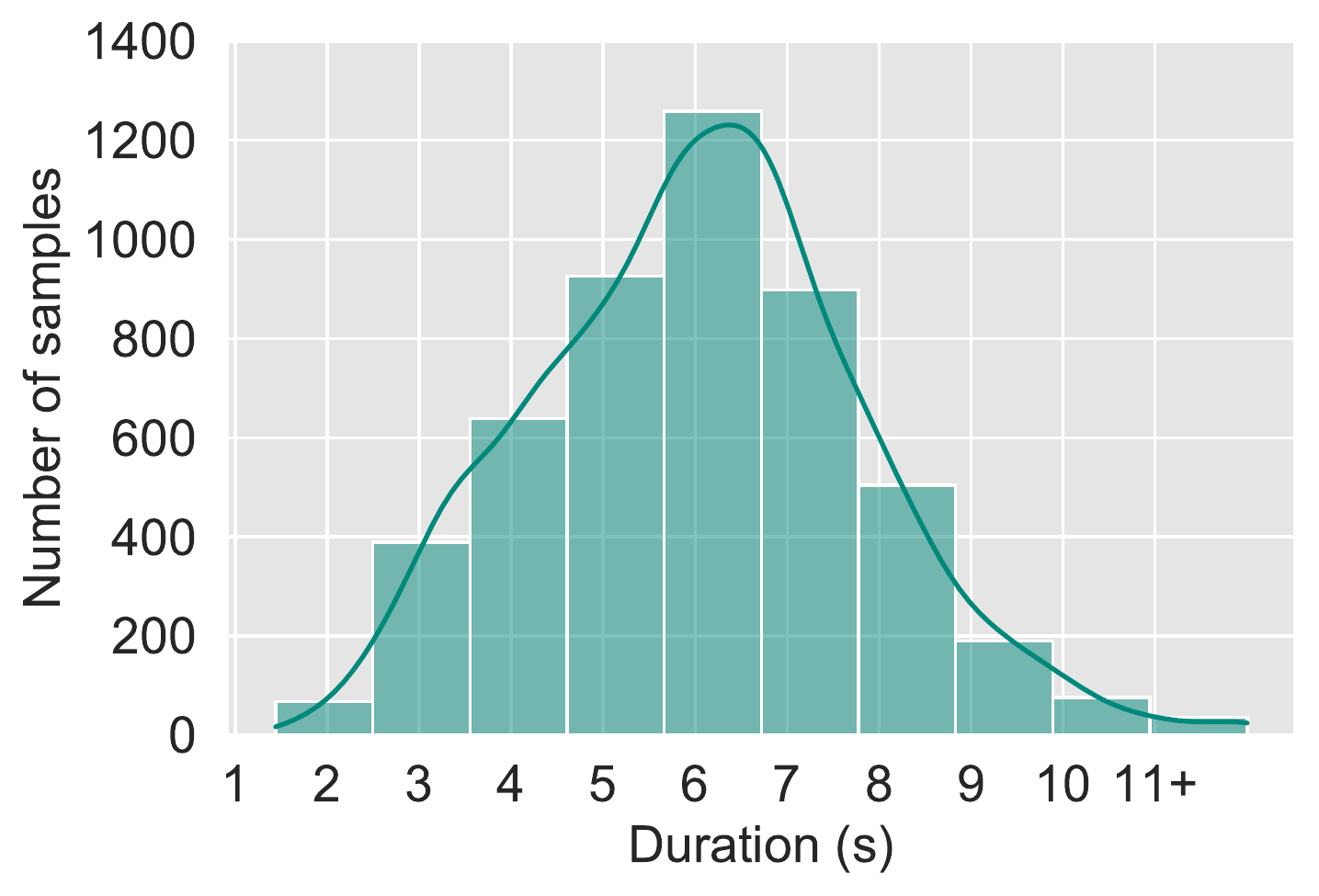}
    \caption{Duration distribution of the recorded data.
    }
    \label{fig:duration}
\end{figure}

We split the data into train, validation, and test sets (Table~\ref{tab:data_split}) by speakers while maintaining a balanced gender distribution for each split. We show the distribution of command length in Figure~\ref{fig:num_chars} and the distribution of sample duration in Figure~\ref{fig:duration}. 

To simulate in-car environments and increase the data scale, we augment each sample from the collected dataset by combining it with 10 different in-car ambient noises that are commonly heard in the daily usage of cars, including \textit{alarm}, \textit{horn}, \textit{background music}, \textit{ignition},  \textit{hail}, \textit{rain}, \textit{windscreen wiper}, \textit{road ambience}, \textit{door opens and closes}, and \textit{people talking}. For each type of noise, we use five variants to increase the diversity and uniformly sample one when applying to the data. The volume of the noise is adjusted by human experts so that the original commands are still recognizable and all the sounds are on the same level of loudness. Therefore, the resulting augmented dataset is 10 times as large as the original clean one and more in line with actual in-car scenarios. As shown in Table~\ref{tab:split_stats}, there are \textit{in-domain} and \textit{out-of-domain} noisy data for different subtasks.

\begin{CJK}{UTF8}{bsmi}
\begin{table}[t]
\centering
\resizebox{\linewidth}{!}{ 
\begin{tabular}{l|l}
\toprule
Command Category &
  Command Examples \\ \midrule\midrule
Navigation &
  \begin{tabular}[c]{@{}l@{}}1. 導航唔該車我去香港科技大學。\\ (Please navigate me to HKUST, thanks.)\\  2. 導航唔該車我去香港藝術館。\\ (Please navigate me to the HK art museum, thanks.)\\ 3. 邊條路可以最快去到維多利亞港?\\ (What is the fastest way to get to Victoria Harbour?)\end{tabular} \\\midrule
Music Playing &
  \begin{tabular}[c]{@{}l@{}}1. 播放張國榮的我。\\ (Play Leslie Cheung's me)\\ 2. 播放Beyond的海闊天空。\\ (Play Beyond's Sea and Sky)\\ 3. 五月天啲歌唔錯，播嚟聽下。\\ (Mayday's song is not bad, let's listen to it)\end{tabular} \\\midrule
Weather Inquiry &
  \begin{tabular}[c]{@{}l@{}}1. 明天天氣如何？\\ (What's the forecast for tomorrow?)\\ 2. 今天晚上天氣如何？\\ (What's the forecast for tonight?)\\ 3. 唔該講下今天天氣點啊?\\ (Tell me the weather of today, thanks)\end{tabular} \\\midrule
Others &
  \begin{tabular}[c]{@{}l@{}}1. 開大啲個冷氣\\ (Turn up the air conditioning)\\ 2. 校細聲啲\\ (Turn down the sound)\\ 3. 開大啲個窗\\ (Open the windows wider）\end{tabular} \\ \bottomrule
\end{tabular}
}
\caption{Examples of commands in different categories with the English translation.}
\label{tab:commands_template}
\end{table}
\end{CJK}

\subsection{Challenge Tasks}

We invite participants to model multimodal automatic speech recognition solutions (ASR) on the CI-AVSR dataset to enable better user experience for in-car smart assistant system. Specifically, we propose three tracks to address different difficulty of challenges in the multimodal ASR as shown in Table 2. The evaluation script will be also provided along with the dataset. We will mark any submission which applies augmentation with an \textbf{asterisk symbol (*)} to distinct the evaluation performance with other non-augmentation approaches.

\subsection{Subtask 1: Clean Speech Cantonese Multimodal ASR}

\begin{table}[]
\centering
\resizebox{\linewidth}{!}{
\begin{tabular}{ll|ccc}
\toprule
\multicolumn{2}{l|}{Data Type} & Train  & Validation   & Test  \\ \midrule \midrule
\multicolumn{2}{l|}{Clean}             & 3,585  & 600   & 799   \\ \midrule
\multirow{2}{*}{Noisy} & \textit{in-domain}     & 17,925 & 3,000 & 3,995 \\
                       & \textit{out-of-domain} & - & - & 3,995 \\ \bottomrule
\end{tabular}
}
\caption{Number of data samples in different data splits. There are five \textit{in-domain} noises 
for the subtask 2 and five \textit{out-of-domain} noises 
for the subtask 3.}
\label{tab:split_stats}
\vspace{-10pt}
\end{table}

In subtask 1, we simulate an ideal situation for an ASR system where clean video and audio data is provided to participants. The clean multimodal data consists of the clean audio data without any background noise and the video of the front facing view of the speaker. Participants are invited to build a multimodal Cantonese ASR model using only the \textbf{train-clean} and \textbf{valid-clean} splits as the training and validation set, respectively. We as organizers will evaluate their trained models on the \textbf{test-clean} split. The output of the ASR models are graphemes of the corresponding speech command which vocabulary consists of 284 written Cantonese characters and 26 lower-cased English alphabet. We will provide the complete vocabulary along with the dataset.

\subsection{Subtask 2: Cantonese Multimodal ASR with In-Domain Ambient Noises}

In the subtask 2, we move closer to a real scenario of in-car multimodal ASR setting, where the given speech command might be intervened by some noises such as the sounds of the engine combustion, horn, and other ambient noises. In this subtask, we assume that all the ambient noises are known such that additional training data with similar characteristic noises can be generated. 
Specifically, in subtask 2, we focus on evaluating model's performance on in-domain (i.e. seen in the training set) data with five different noises, including windscreen wiper, horn, alarm, engine ignition, and door noises.
Participants are allowed to train their models on the \textbf{train-clean} and \textbf{train-noisy (in-domain)} sets, and validate their models on both of the \textbf{valid-clean} and \textbf{valid-noisy (in-domain)} sets. We will score their results on the \textbf{test-noisy (in-domain)}.


\subsection{Subtask 3: Cantonese Multimodal ASR with Out-of-Domain Ambient Noises}

In subtask 3, we move even further to the real scenario of in-car multimodal ASR setting, where we assume that the characteristics of noises are unknown and only appear during the inference time. 
More specifically, we focus on evaluating the generalization ability of models to see whether they can still perform well on out-of-domain (i.e. unseen in the training data) noisy data, which includes five various noises (background music, rain, hail, people talking, road ambience).
Participants are allowed to train their models on the \textbf{train-clean} and \textbf{train-noisy (in-domain)} sets, and validate their models on the \textbf{valid-clean} and \textbf{valid-noisy (in-domain)} sets. We will evaluate their models on the \textbf{test-noisy (out-of-domain)} set. 

\subsection{Evaluation}
We will provide a test set for each subtask and the participants are invited to submit the generation results for evaluating their models's performance. We will use character error rate (CER) and mixed error rate (MER) as evaluation metrics. In detail, the commands in the test set will be divided into two groups. For the commands with all Cantonese characters we use CER as the evaluation metrics, while for code-switched commands which include Latin alphabets, we apply MER for them.

\section{Task Organizers}

\begin{itemize}
    \item \small{Wenliang Dai, wdaiai@connect.ust.hk}
    \item \small{Samuel Cahyawijaya, scahyawijaya@connect.ust.hk}
    \item \small{Tiezheng Yu, tyuah@connect.ust.hk}
    \item \small{Elham J. Barezi, eebarezi@connect.ust.hk}
\end{itemize}


\bibliography{anthology,custom}
\bibliographystyle{acl_natbib}




\end{document}